\documentclass[tinyml, table,xcdraw]{acmart}

\AtBeginDocument{%
  \providecommand\BibTeX{{%
    \normalfont B\kern-0.5em{\scshape i\kern-0.25em b}\kern-0.8em\TeX}}}

\setcopyright{rightsretained}
\copyrightyear{2024}
\acmYear{2024}

\usepackage{multirow}
\usepackage{graphicx}
\usepackage{enumitem}
\usepackage{cleveref}
\usepackage{subfig}
\usepackage{graphicx}
\usepackage{xcolor}

\newcommand{\sys}{TinyVQA}
\begin{document}

\title{TinyVQA: Compact Multimodal Deep Neural Network for\\ Visual Question Answering on Resource-Constrained Hardware}

\author{Hasib-Al Rashid}
\authornote{Both authors contributed equally to this research.}
\affiliation{%
  \institution{University of Maryland, Baltimore County}
  \state{Maryland}
  \country{USA}
  \postcode{21250}
}
\email{hrashid1@umbc.edu}

\author{Argho Sarkar}
\authornotemark[1]
\affiliation{
  \institution{University of Maryland, Baltimore County}
  \streetaddress{1000 Hilltop Circle}
  \state{Maryland}
  \country{USA}
  \postcode{21250}
}
\email{asarkar2@umbc.edu}

\author{Aryya Gangopadhyay}
\affiliation{
  \institution{University of Maryland, Baltimore County}
  \streetaddress{1000 Hilltop Circle}
  \state{Maryland}
  \country{USA}
  \postcode{21250}
}

\author{Maryam Rahnemoonfar}
\affiliation{
  \institution{Lehigh University}
  \state{Pennsylvania}
  \country{USA}
}

\author{Tinoosh Mohsenin}
\affiliation{
  \institution{Johns Hopkins University}
  \state{Maryland}
  \country{USA}
  \postcode{21250}
}

\begin{abstract}
Traditional machine learning models often require powerful hardware, making them unsuitable for deployment on resource-limited devices. Tiny Machine Learning (tinyML) has emerged as a promising approach for running machine learning models on these devices, but integrating multiple data modalities into tinyML models still remains a challenge due to increased complexity, latency, and power consumption. This paper proposes TinyVQA, a novel multimodal deep neural network for visual question answering tasks that can be deployed on resource-constrained tinyML hardware. TinyVQA leverages a supervised attention-based model to learn how to answer questions about images using both vision and language modalities. Distilled knowledge from the supervised attention-based VQA model trains the memory aware compact \sys{} model and low bit-width quantization technique is employed to further compress the model for deployment on tinyML devices. The TinyVQA model was evaluated on the FloodNet dataset, which is used for post-disaster damage assessment. The compact model achieved an accuracy of \textbf{79.5\%}, demonstrating the effectiveness of TinyVQA for real-world applications. Additionally, the model was deployed on a Crazyflie 2.0 drone, equipped with an AI deck and GAP8 microprocessor. The TinyVQA model achieved low latencies of \textbf{56 ms} and consumes \textbf{693 mW} power while deployed on the tiny drone, showcasing its suitability for resource-constrained embedded systems.
\end{abstract}



\maketitle
\vspace{-2ex}
\section{Introduction}


\begin{figure*}
\begin{center}
\includegraphics[width=0.8\textwidth]{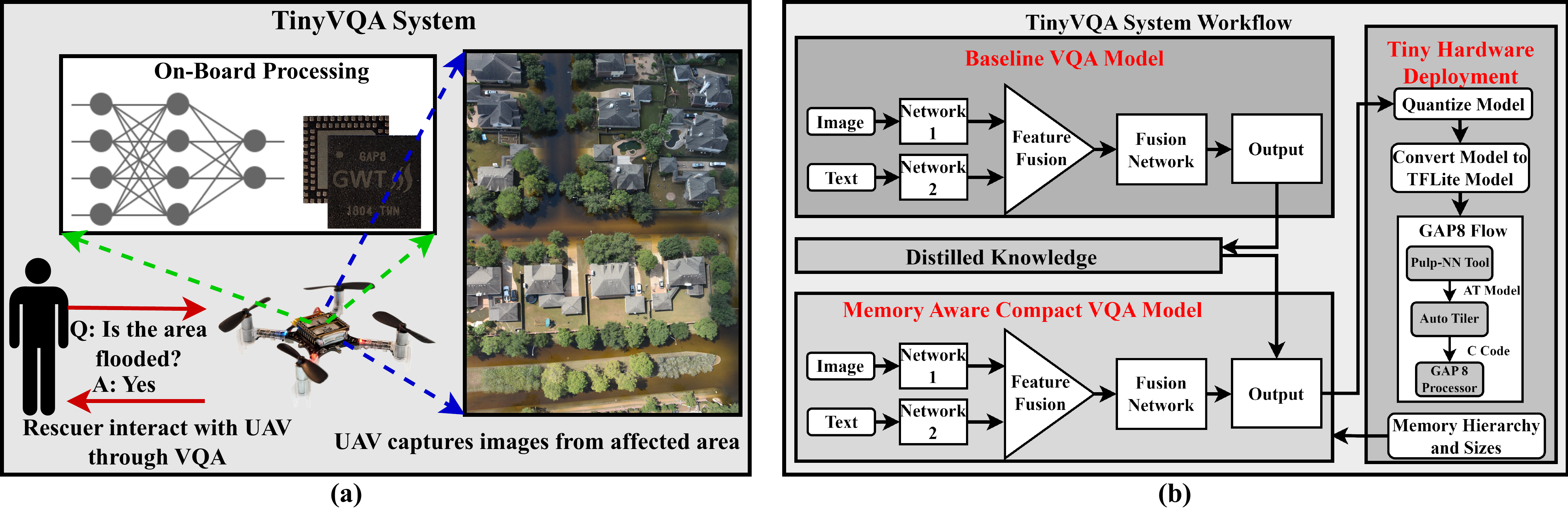}
\end{center}
\vspace{-2ex}
\caption{\small (a) Highlevel overview of proposed \sys{} system. Rescuer can acquire effective information about the affected area by asking questions when a drone coupled with a VQA system captures images
from the hurricane-stricken area from a high altitude. (b) The flow diagram of proposed \sys{} system. Proposed \sys{} is the sequential combination of the steps shown in the diagram.
\label{highlevel1}}
\vspace{-3ex}
\end{figure*}
\vspace{1ex}

Tiny Machine Learning (tinyML) is rapidly transforming how we deploy machine learning models to resource-constrained devices at the edge, particularly in remote or disaster-stricken areas. Its minimal power consumption and independence from internet connectivity make it ideal for scenarios where real-time insights are crucial, but resources are limited. While traditionally focused on analyzing single data modalities, tinyML now faces the exciting challenge of integrating multimodal deep neural networks (M-DNNs) \cite{rashid2022tinym2net, rashid2023tinym2netv2, rashid2, rashid2023hac, ovi2021aris, rashid2022coughnet, ovi2022towards}. These networks leverage data from multiple sources, such as text, audio, and image, to provide richer insights and more robust predictions for tasks like Visual Question Answering (VQA). VQA technology presents a powerful solution for many computer vision tasks by combining vision and language modalities to extract high-level scene information from images. Unlike traditional computer vision tasks, VQA's ability to answer questions about images using natural language streamlines the decision-making process, providing critical insights for immediate and long-term response efforts \cite{vqaaid, uavvqg, argho1, argho2}.

However, implementing M-DNNs (i.e. VQA) on resource constrained tiny devices presents significant challenges. Their high computational complexity and memory footprint, coupled with the intricacies of optimizing them for resource-constrained platforms, pose formidable barriers. Fortunately, recent advancements in model compression techniques, including parameter pruning, knowledge distillation, and quantization, offer promising solutions \cite{lin2020mcunet, banbury2021micronets, mazumder2021survey, saeid, ISQED20-hasib, jetc2020-hasib, Khan2023} . These techniques have been effective in reducing the size and complexity of unimodal models, paving the way for adapting M-DNNs for edge deployment. MobiVQA \cite{mobivqa} proposed on-device VQA, focusing on early exit and selective processing and optimizing existing VQA models for mobile devices. However, their implementation is not suitable for tinyML hardware deployment.

The integration of TinyML and VQA holds immense potential to revolutionize disaster preparedness, response, and recovery efforts. This transformative technology empowers communities with real-time, on-site intelligence, ultimately saving lives and minimizing the impact of catastrophic events. However, current state-of-the-art VQA models rely heavily on cloud servers and GPUs due to their resource-intensive nature, rendering them impractical in disaster scenarios with limited resources or compromised infrastructure. As the integration of TinyML with multimodal Visual Question Answering (VQA) is a burgeoning area of research, it presents an untapped opportunity for innovation in disaster management. The potential for developing refined M-DNN models tailored for resource-limited platforms like TinyML heralds a promising future in effectively mitigating the impacts of natural disasters.

In this paper, we introduce \emph{\sys{}}, a compact multimodal deep neural network specifically designed for Visual Question Answering (VQA) tasks, tailored for deployment on resource-constrained TinyML hardware. Highlevel overview of the proposed \sys{} is presented in the figure \ref{highlevel1} (a). To the best of our knowledge, this represents the first attempt to deploy a VQA task on extremely resource-limited hardware. The main contributions of this paper can be summarized as follows:
\vspace{-1ex}
\begin{itemize}[leftmargin=0.25in]
 \item We propose \emph \sys{}, a novel multimodal (vision and language based) deep neural network for visual question answering task, deployable on resource-constrained tinyML hardware.
 \item We designed supervised
    attention-based visual question answering framework, distilled its knowledge to train our compact, memory-aware \sys{} model.
 \item We evaluated our \sys{} with FloodNet \cite{floodnet, sarkar2023sam} dataset, which is used for post-disaster damage assessment purposes, conveying the efficacy of tinyML during disaster management.
 \item We deployed our compact \sys{} model on the resource-constrained Crazyflie 2.0 drone, equipped with an AI-deck, which operates using the GAP8 microprocessor. We conducted an analysis of the onboard latency and power consumption for the proposed \sys{} architecture.
\vspace{-2ex}
\end{itemize}

\section{\protect\sys{} Model Architecture}
The figure \ref{highlevel1} (b) delineates the workflow of the TinyVQA System, which consists of two main components: the Baseline VQA Model and the Memory Aware Compact VQA Model.
\vspace{-2ex}
\subsection{Baseline VQA Model Design}

\begin{figure*}
\begin{center}
\includegraphics[width=0.8\textwidth]{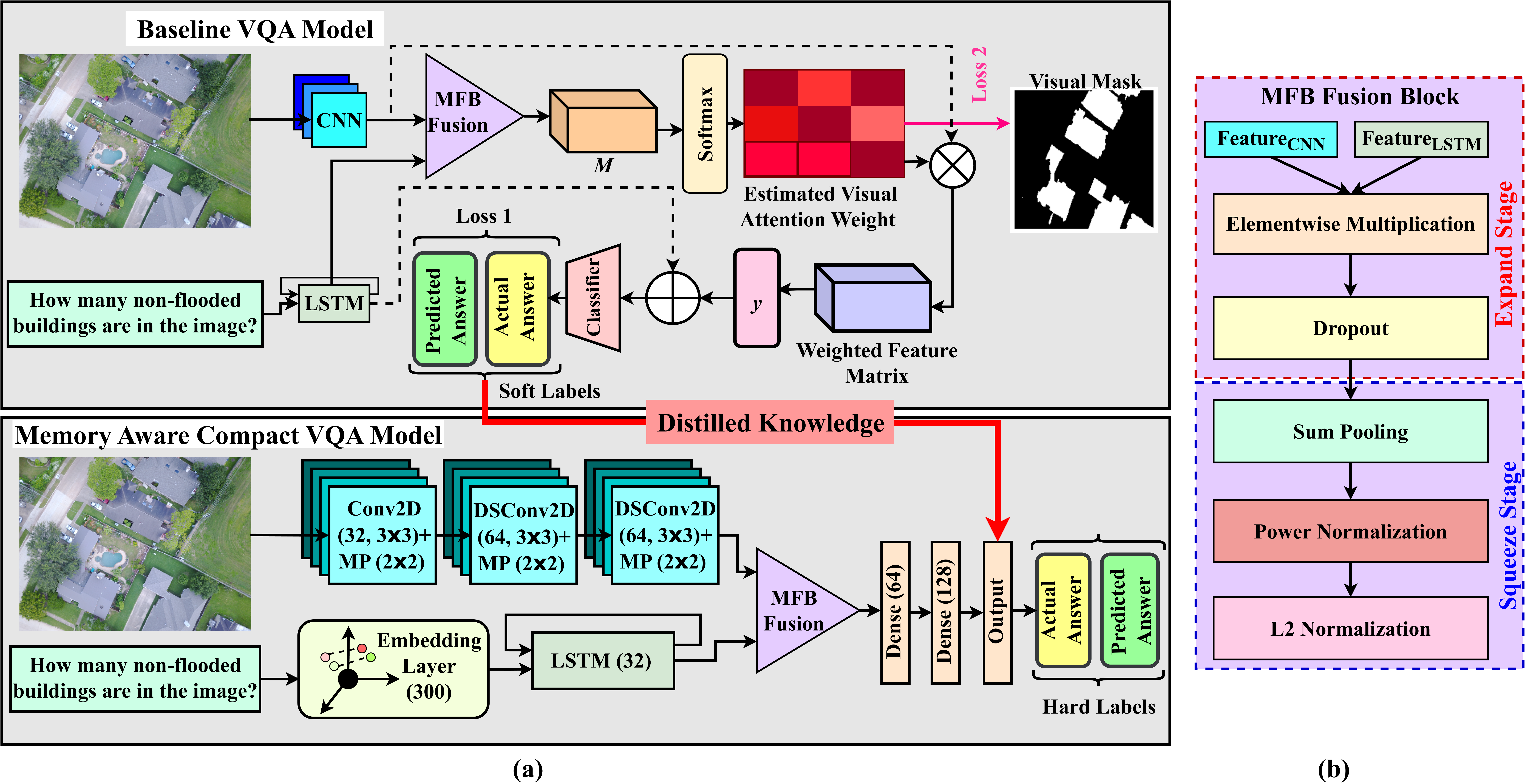}
\end{center}
\vspace{-2ex}
\caption{\small (a) Overview of our proposed \sys{} model where the baseline VQA model uses VGG-16 and a one-layer LSTM to obtain the image feature matrix and question feature, respectively. We then consider MFB pooling to obtain a fine-grained multimodal representation. A softmax function is applied to that joint representation to estimate attention weights from the images for given questions. Finally, we calculate two loss functions: one minimizes the distance between
the visual mask and the estimated visual attention weight, and the other minimizes the loss between the ground-truth answer and the predicted answer from
the VQA classifier. Memory aware compact VQA model is designed with 3 CNN layers and 1 LSTM layer for each of the image and text modality feature extraction. Distilled knowledge is used from the baseline model to have the final result. (b) Detailed structure of the MFB Fusion block.
\label{model}}
\vspace{-3ex}
\end{figure*}

We designed a vision-based question-answering model following the approach mentioned in the previous work \cite{sarkar2023sam}. This model allows decision-makers to pose image-specific questions in natural language, enabling the model to generate answers derived from visual content. We adopted an enhanced visual attention mechanism by introducing an auxiliary visual mask, which refines the model's estimated attention weights. This augmentation is designed to work in harmony with the cross-entropy loss used for classification, thereby not only improving the model’s accuracy but also its interpretability. The architecture of the model is structured into four distinct stages: Visual Feature Extraction, Textual Feature Extraction, Fine-Grained Fusion, and Classification, with the comprehensive architecture of this baseline model illustrated in Figure \ref{model}.
\vspace{-2ex}

\par In the visual feature extraction stage, the process involves feeding an image into a Convolutional Neural Network (CNN), specifically VGG-16. This step aims to extract the grid feature matrix from the final convolutional layer of the CNN model. When the image $I$ has an input size of $224 \times 224$, the CNN model generates a grid feature matrix $f_I$ with dimensions $14 \times 14 \times 1024$. This matrix extracts the relevant visual features from the image, forming a semantic visual representation that serves as a foundation for subsequent stages in the processing pipeline. 
\begin{equation}
f_{\mathcal{I}} = VGG({\mathcal{I}}) \in \mathbb{R}^{h \times w \times k}
\end{equation}

Here, $h$, $w$, and $k$ represent the height, width, and number of channels of the image feature matrix $f_I$, respectively.

Within the textual feature extraction stage, a two-layer Long Short-Term Memory (LSTM) architecture is employed to extract the $1024$-dimensional semantic question feature $f_T$ from the input question $Q$. This specific feature is derived from the last cell of the final layer within the LSTM model. Before being input into the LSTM layer, all questions are tokenization and padding with $0$ to guarantee a consistent length.
\begin{equation}
f_{\mathcal{T}}=LSTM (\mathcal{Q}) \in \mathbb{R}^{k}
\end{equation}

\par Within the fine-grained fusion stage, the Multi-Modal Factorized Bilinear (MFB) pooling technique is employed to integrate each grid-feature vector from an image feature matrix with the semantic textual feature extracted from the corresponding question. This fusion process is structured into two stages, the expand stage and the squeeze stage. In the expanding process, the image grid vector and question feature vectors undergo element-wise multiplication, followed by the introduction of a dropout layer. The subsequent squeezing step involves sum pooling, followed by power and $l_2$ normalization layers. The resulting matrix is then subjected to the Softmax function to compute the spatial visual attention weight. Notably, an auxiliary visual mask, derived from ground-truth visual attention based on a semantic segmented image, is employed to guide the estimated spatial attention weight in this phase.

\par Following the estimation of spatial attention weight, it is multiplied with the image feature matrix and summed channel-wise. The resultant summed vector is subsequently added to the question vector, forming the input for the classification layer responsible for generating the final prediction. This comprehensive fusion strategy enhances the model's ability to capture high-level relationships between visual and textual features for accurate and meaningful predictions. 

\subsection{Memory-Aware Compact VQA Model Design}
We designed a novel compact model introducing memory-aware model compression of the large, complex M-DNN model. Our proposed memory-aware compression technique for Multimodal Deep Neural Networks (M-DNNs) employs off-the-shelf knowledge distillation and quantization to minimize the memory footprint of M-DNNs while preserving their accuracy. This method involves training a compact student M-DNN to emulate a more complex teacher model, aiming to fit the student model within the constrained on-chip memories (L1 and L2) of tiny processors. The focus is on striking a balance between model size reduction and performance retention, making it suitable for deployment on resource-limited devices. To this extent, we considered several factors in memory-aware model compression:
\par \textbf{Memory Hierarchy of the Deployment Hardware:} Memory aware model compression in \sys{} takes into account the memory hierarchy of the target hardware platform, such as on-chip SRAM and off-chip DRAM, to ensure that the compressed model can be effectively stored and executed within the available lower level of memory hierarchies for faster and more efficient deployment.
\par \textbf{Model Compression Techniques:} To compress the large multimodal neural network models, \sys{} used off-the-shelf model compression  such as: Knowledge Distillation, Uniform 8-bit Quantization and Compact Network architecture design for inference.

Knowledge Distillation involves training a smaller student model to mimic the behavior of a larger, more accurate teacher model. The student model is designed to be smaller than the teacher model, with fewer layers, parameters, or connections, to reduce memory requirements for deployment on memory-constrained devices. Various distillation techniques, such as soft targets, attention transfer, or feature map matching, can be used for transferring knowledge from the teacher to the student model. We have chosen to use soft targets for this purpose. Soft targets are generated by the larger model and used as training labels for the smaller model. These are obtained by applying a temperature scaling factor, \(T\), to the output probabilities of the larger model, which smooths the peaks in the distribution, making it more spread out. The soft targets, denoted as \(q_i\), are defined by the following equation:

\begin{equation}
q_i = \frac{\exp(z_i/T)}{\sum_j \exp(z_j/T)}
\end{equation}

where \(z_i\) is the logit (unnormalized log-probability) output of the larger model for class \(i\). The temperature scaling factor, \(T\), determines the "softness" of the targets, with higher values leading to softer targets. The smaller model is then trained to minimize the Kullback-Leibler (KL) divergence between its output probabilities, \(p_i'\), and the soft targets, \(q_i\). This is expressed by the following loss function:

\begin{equation}
\mathcal{L} = \sum_i q_i \log\frac{q_i}{p_i'}
\end{equation}

where \(p_i'\) represents the output probability of the smaller model for class \(i\). This KL divergence loss function encourages the smaller model to learn a probability distribution similar to that of the larger model.

Designing compact and efficient network architectures can help create models with smaller memory footprints without sacrificing accuracy. We have used depthwise separable convolution layers in stead of regular convolution layers to further compress the student model without significant loss of accuracy. Reducing the bit-width of the model parameters and activations can significantly reduce the memory footprint of the model. We quantized our models with Tensorflow Lite (tf-lite) post-training quantization adopting full integer quantization. This method quantizes both the weights and activations to 8-bit integers, resulting in a model that performs only integer arithmetic. The hyper parameters of the compact model are fixed empirically.

\section{\protect\sys{} Evaluation}
\label{eve}
\subsection{Dataset Description}

\par In this study, we consider the \textit{FloodNet-VQA} dataset \cite{floodnet} for the experiment. The data collection process occurred in the aftermath of Hurricane Harvey, a devastating Category 4 hurricane that struck Texas and Louisiana in August 2017, resulting in widespread flooding and a tragic loss of over 100 lives. Leveraging the unmanned aerial vehicle (UAV) platform, images and videos of the affected regions were captured. Specifically, DJI Mavic Pro quadcopters were employed for this data collection initiative. Numerous flights were conducted to cover primarily Ford Bend County, Texas, and other directly affected areas. The dataset comprises a total of $2348$ images and $10,480$ question-answer (QA) pairs. Within the dataset, there are 7 distinct question categories, including \textit{Simple Counting}, \textit{Complex Counting}, \textit{Road Condition Recognition}, \textit{Density Estimation}, \textit{Risk Assessment}, \textit{Building Condition Recognition}, and \textit{Entire Image Condition}. These diverse question categories are crucial for a comprehensive understanding of the damaged scenario, ultimately contributing to the effectiveness of rescue missions. Notably, the dataset encompasses questions with varying complexities, with the longest question containing $11$ words.

\subsection{Evaluation Results and Analysis}

\begin{figure}
\begin{center}
\includegraphics[width=0.4\textwidth]{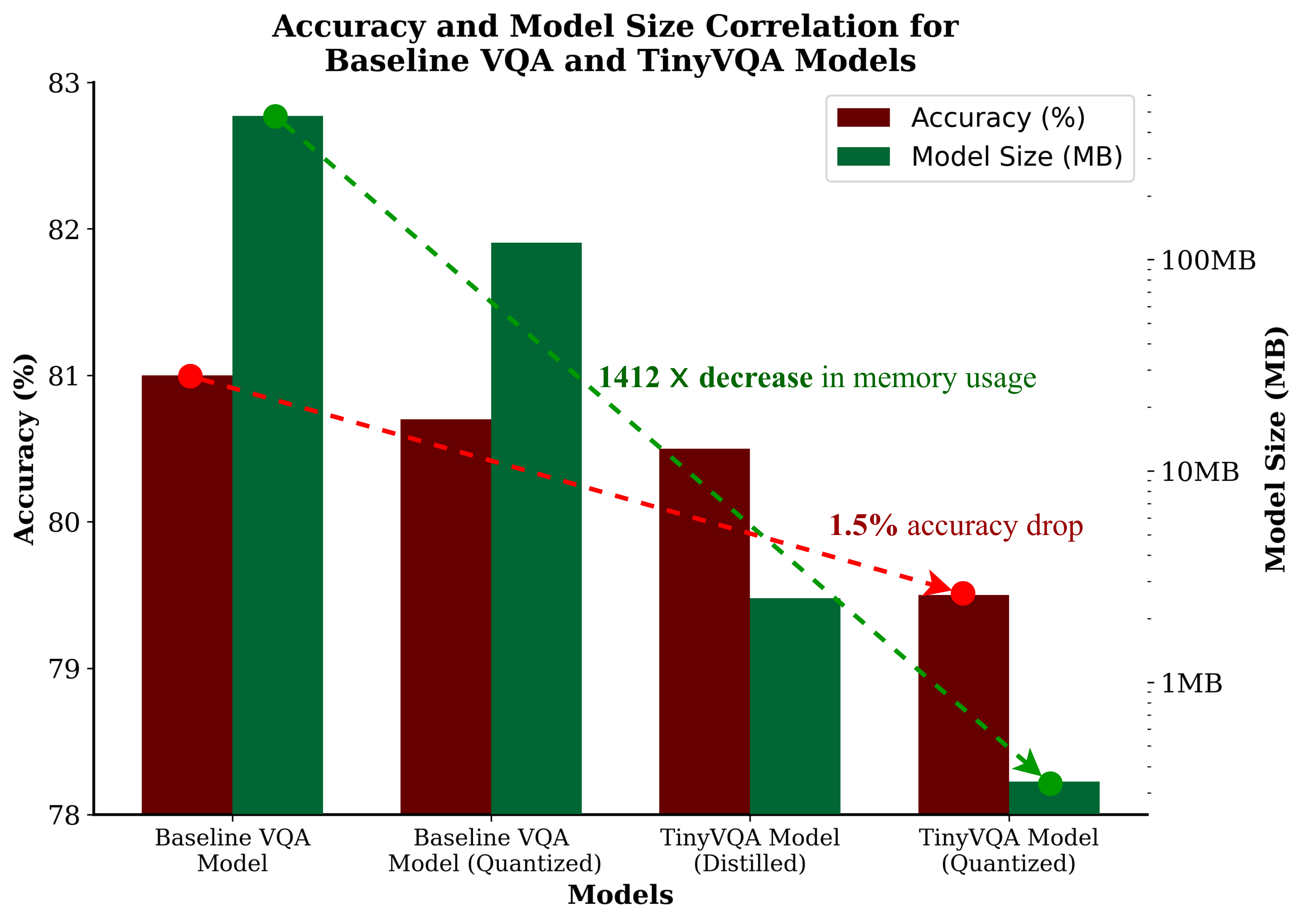}
\end{center}
\vspace{-2ex}
\caption{\small Accuracy and Model Size Correlation for Baseline VQA and TinyVQA for FloodNet \cite{floodnet} dataset. Baseline model achieved 81\% accuracy with 479 MB model size whereas final TinyVQA model achieved 79.5\% accuracy with 339 KB model size.
\label{results}}
\vspace{-3ex}
\end{figure}

Figure \ref{results} presents \sys{} evaluation results for baseline VQA and TinyVQA on FloodNet \cite{floodnet} dataset. The baseline model achieved 81\% accuracy with 479 MB model size whereas final TinyVQA model achieved 79.5\% accuracy with 339 KB model size. The figure \ref{results} presents a comparative analysis of the Baseline VQA and TinyVQA Models, focusing on the trade-off between accuracy and model size. The Baseline VQA Model, when both distilled and quantized to TinyVQA model, shows a 100\% decrease in memory usage with a corresponding 1.5\% drop in accuracy which exhibit a huge reduction in model size, indicating an optimization for memory-constrained deployments. The downward trajectory of the dashed line highlights the inverse relationship between model size and accuracy, emphasizing efficiency gains in the TinyVQA Models at a slight cost to accuracy.

\begin{figure}
\begin{center}
\includegraphics[width=0.42\textwidth]{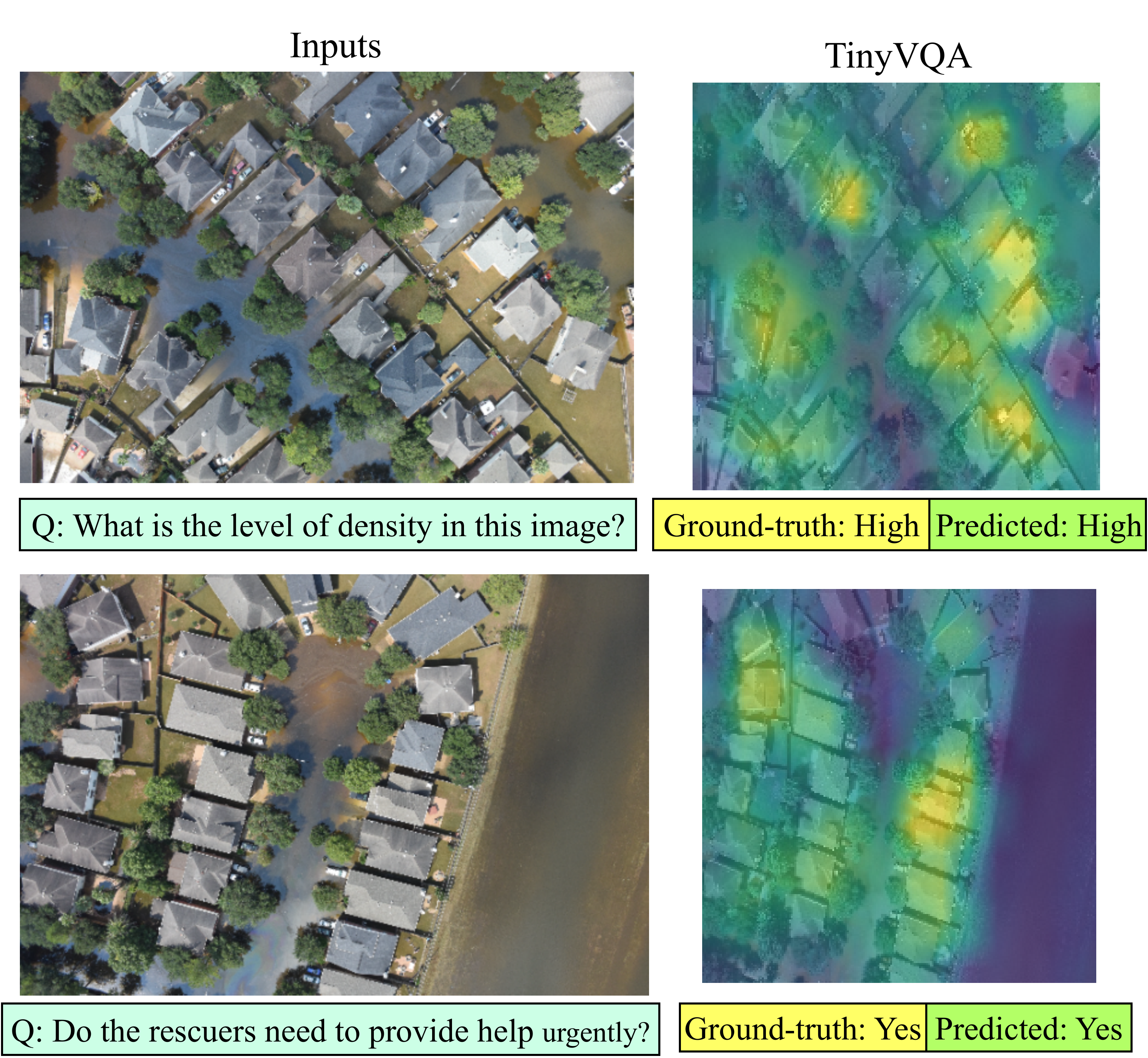}
\end{center}
\vspace{-2ex}
\caption{\small Derived visual attentions for given questions from \sys{} model. The yellowish tone in the image denotes higher attention weight. Attention learned with visual supervision (the last column) emphasizes the relevant image portions (buildings and roads in this case) to address the questions from the top and bottom images, respectively.
\label{fig:v_attention}}
\vspace{-3ex}
\end{figure}

Figure \ref{fig:v_attention} represents the qualitative results from TinyVQA model. This figure shows that derived visual attention from the TinyVQA model focuses on the relevant image portions depending on the questions. These qualitative results prove the trustworthiness of the predictions from our model. 

\vspace{-3ex}

\begin{figure*}
\begin{center}
\includegraphics[width=0.9\textwidth]{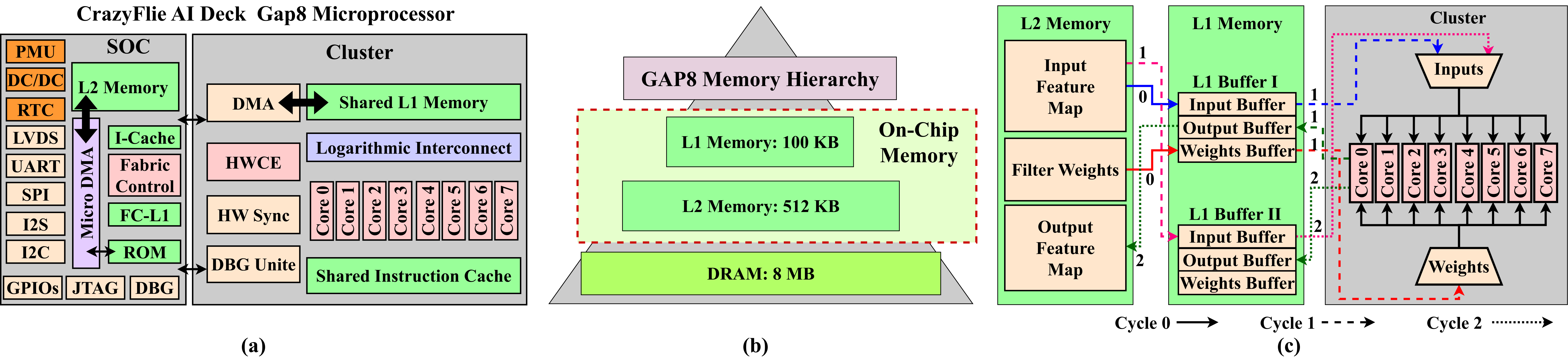}
\end{center}
\vspace{-2ex}
\caption{\small (a) Detailed block diagram of Crazyflie AI-deck powered by GAP8 microprocessor (b) Memory Hierarchy for GAP8 microprocessor. GAP 8 microprocessor has L1 Memory of 100 KB (80 KB shared in compute engine + 20 KB for low power MCU.), L2 memory of 512 KB and 8MB of DRAM (c) TinyVQA flow; Left: the DMA manages L2 -L1 communication using double-buffering. Right: the cluster executes PULP-NN on tile stored in one of the L1 buffers.
\label{hardware}}
\vspace{-3ex}
\end{figure*}
\vspace{1ex}

\section{\protect\sys{} Deployment on Resource Constrained Hardware}
\subsection{Hardware Architecture}
Crazyflie 2.0 tiny drone equipped with the AI-deck, powered by the GAP8 RISC-V microprocessor \cite{gap81, gap82}, has become very popular to deploy tinyML models for UAVs \cite{navardi2023mlae2, manjunath2023reprohrl, navardi2023metae2rl, humes2023squeezed}. We have also chosen Crazyflie 2.0 as the deployment platform testbed for our tinyML model. The GAP8 microprocessor emerges as a powerful contender for edge AI applications, wielding a multi-core RISC-V architecture optimized for parallel processing and hardware-accelerated deep learning. A single RISC-V core, the Fabric Controller (FC), acts as the conductor, orchestrating operations, scheduling tasks, and managing peripheral connections. Eight additional RISC-V cores in the Compute Cluster work in tandem, tackling compute-intensive tasks like image and audio processing. For CNNs, the dedicated Hardware Convolution Engine (HWCE) steps in, boosting AI performance without compromising energy efficiency. A layered memory hierarchy ensures smooth data flow: private L1 caches for each core and a shared L2 pool of 512 KB offer ample storage for data and code. Independent DMA units, the Micro-DMA and Cluster-DMA, handle complex I/O tasks and data transfers between memory levels, further fueling efficient operation. Extensive peripheral interfaces provide seamless integration with sensors, cameras, and other external devices, enabling versatile edge applications. Moreover, GAP8 excels in power management, utilizing power-gating for idle components and dynamic voltage/frequency scaling to adapt to varying workloads. This intelligent power handling, coupled with integrated DC/DC regulators and clock generators, allows for rapid reconfiguration and maximizes energy efficiency. In conclusion, with its focus on parallel processing, dedicated AI acceleration, and robust memory management, the GAP8 architecture establishes itself as a compelling platform for realizing AI-powered edge applications, even under tight energy constraints. Figure \ref{hardware} (a) and (b) shows the memory hierarchy and detailed block diagram for GAP8 microprocessor.
\vspace{-2ex}
\subsection{Hardware Implementation Details}
The GAPFlow toolchain, comprising PULP-based Neural Network Tool (NNTooL) and AutoTiler, played a pivotal role in deploying the DNN onto the GAP8 architecture (Fig. \ref{highlevel1}(b)) \cite{puplnn, pulpnn2}. NNTooL meticulously adapted the DNN architecture to ensure compatibility with AutoTiler and meticulously transformed weights for optimal execution on GAP8. AutoTiler then took the reins, algorithmically optimizing memory layout and generating GAP8-compatible C code, streamlining the deployment process. Despite the toolchain's automation, manual intervention was occasionally required to address DNN-specific memory requirements. This included adjusting maximum stack sizes and fine-tuning heap space allocation to prevent potential heap overflows, data corruption, and stack-related issues. Notably, AutoTiler's default behavior of allocating the entirety of L1 and L2 memory could lead to such challenges. Further considerations involved the GAP8's Real-Time Operating System (RTOS), which pre-allocates heap memory, potentially reducing available space for DNN operations, necessitating careful memory management.

The GAP8 architecture presents a promising platform for efficient deployment of multi-modal models at the edge, thanks to its unique blend of hardware acceleration and parallel processing capabilities. To fully leverage these features, we employed a strategic partitioning of model components:

\par \textbf{1. Hardware Acceleration:} The Hardware Convolution Engine (HWCE) was tasked with handling computationally intensive convolutional layers, capitalizing on its dedicated hardware optimization for accelerated execution. The remaining layers, including fully connected layers, activation functions, dropout layers, and the LSTM layer, were executed by the Compute Cluster, comprising eight RISC-V cores operating in parallel. This division of labor effectively harnessed the strengths of each processing unit.

\par \textbf{2. Data Flow and Memory Optimization:} Image data was strategically stored in L2 memory for direct accessibility by the HWCE, streamlining convolutional processing. Intermediate feature maps, generated during model execution, were cached in L1 memory for rapid reuse, optimizing data transfer. Textual input processing was assigned to the RISC-V cores, potentially utilizing general-purpose matrix multiplication libraries for embedding and LSTM computations. To address model size constraints imposed by GAP8's L2 memory capacity, model quantization was employed, reducing data precision to 8-bit integers and shrinking the model's memory footprint. Additionally, the Cluster-DMA unit proactively prefetched data from L2 to L1, anticipating upcoming computations and minimizing memory stalls.

\par \textbf{3. Parallelism for Enhanced Performance:} Parallelism emer-ged as a cornerstone of GAP8's performance advantage. The Fabric Controller (FC) core masterfully orchestrated concurrent operations, expertly dividing neural network tasks between the Compute Cluster and HWCE for seamless collaboration. Layer parallelism further amplified efficiency, allowing different model layers to execute simultaneously on separate cores, unlocking significant speed gains. The FC core also demonstrated ingenuity in offloading data preparation and post-processing tasks to idle cores, effectively overlapping them with model computations and streamlining the overall process. Furthermore, the Micro-DMA unit independently managed I/O operations, enabling parallel data transfers and minimizing processing bottlenecks.

Figure \ref{hardware} (b) and (c) elucidates the memory hierarchy and computational flow orchestrated within the GAP8 processor for efficient execution of CNN-LSTM multimodal models. It highlights the interplay between L2 memory, L1 memory, and the processing cluster, demonstrating how data flows seamlessly across these components over three distinct computation cycles. L2 memory serves as the primary repository for the input feature map, filter weights, and post-processed output feature map. L1 memory, strategically divided into two buffers (L1 Buffer I and L1 Buffer II), employs a double-buffering technique to facilitate concurrent data loading and processing. Each buffer is further partitioned into dedicated areas for inputs, outputs, and weights. The processing cluster, comprising seven cores, forms the heart of the computation. It ingests inputs and weights and executes neural network operations in parallel. The figure \ref{hardware} (c) meticulously depicts this processing over three cycles, effectively illustrating the temporal dynamics of the dataflow:

\par \textbf{Cycle 0:} The initial loading of the input feature map from L2 memory to L1 Buffer I is marked by solid lines.
\par \textbf{Cycle 1:} As indicated by dashed lines, L1 Buffer I processes the first data set while L1 Buffer II concurrently commences loading the subsequent set, showcasing a seamless overlap.
\par \textbf{Cycle 2:} Dotted lines visualize the transfer of L1 Buffer I outputs back to L2 memory and the initiation of processing within L1 Buffer II, ensuring uninterrupted operation and eliminating idle time. 

Arrows of varying styles delineate the data movement across cycles, with color-coded arrows (blue, red, and green) corresponding to the distinct paths for inputs, weights, and outputs, respectively.

\subsection{Deployment Results and Analysis}

\begin{figure}
\centerline{\includegraphics[width=0.3\textwidth]{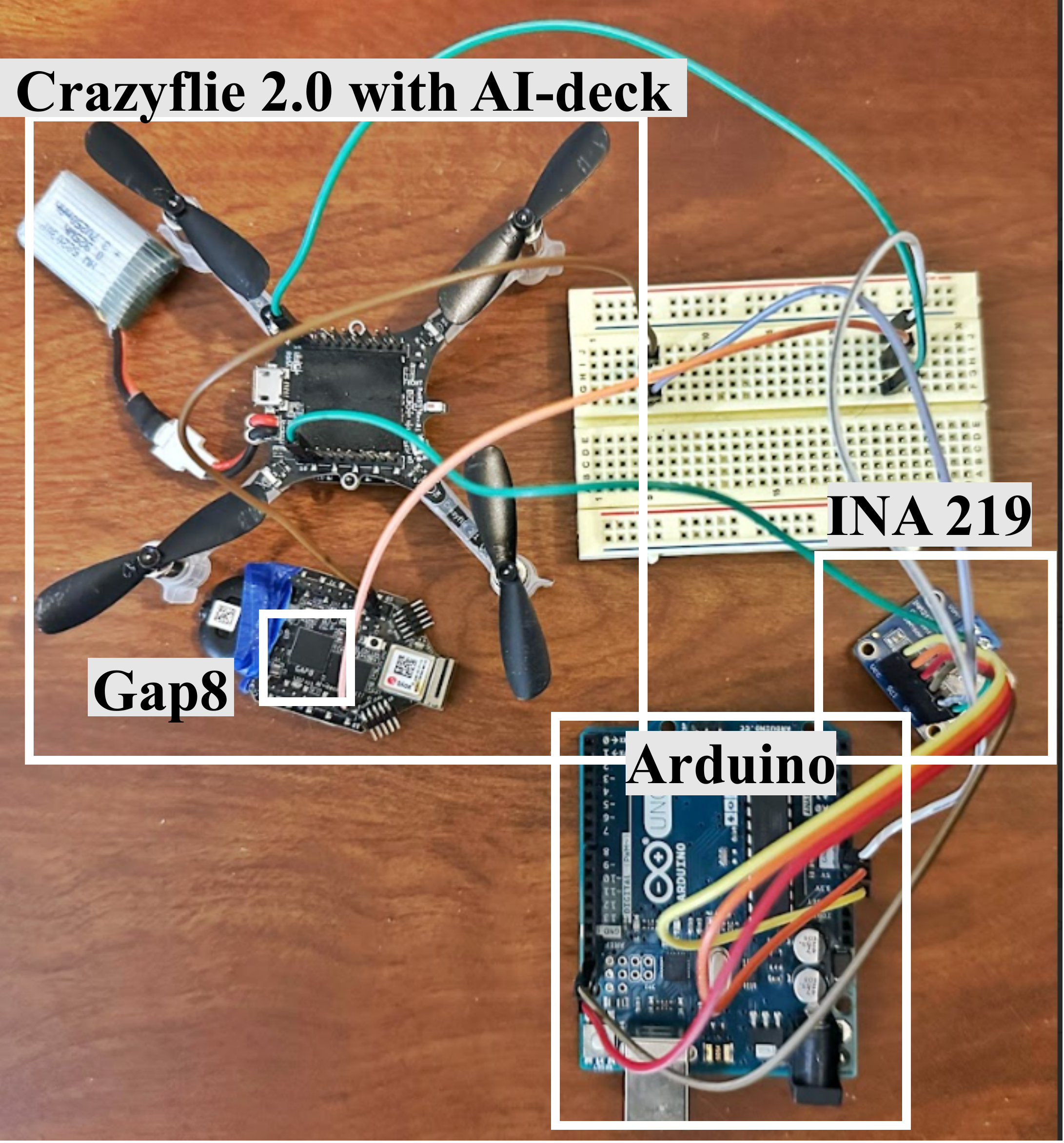}}
\caption{\small Crazyflie Ai-deck board power measurement setup. INA219 and Arduino measure the GAP8 power consumption.}\vspace*{-5pt}
\label{power}
\end{figure}

We deployed the \sys{} model on the GAP8 processor. Table \ref{resource} reports resource utilization of \sys{} for post-disaster damage assessment purposes. \sys{} uses around 49 KB of L1 memory and 312 KB of L2 memory which are 93\% and 78\% of the available L1 and L2 memories. The inference model does not require off-chip DRAM to store its weights and activations which ensures the minimum latency. Figure \ref{power} (a) displays the power measurement setup used in this work for Creazyflie 2.0 with AI-deck, using INA 219 sensor and Arduino board.

Table \ref{comparison} provides comparative analysis where we examine two architectures, proposed \sys{} and MobiVQA \cite{mobivqa}, designed for multimodal (image and text) question answering tasks. TinyVQA is deployed on a GAP8 processor, while MobiVQA, optimized by Pytorch mobile, utilizes an Nvidia TX2 Board.
TinyVQA exhibits lower operating frequency of 175 MHz but achieves significantly lower latency (56 ms vs. 213 ms of MobiVQA), suggesting real-time suitability. Additionally, TinyVQA shines in energy consumption per query (0.2 J vs. 5.60 J), making it ideal for edge deployments. This represents a significant advantage in scenarios where energy efficiency is paramount, such as battery-powered or remote devices.

\begin{table}[]
\caption{\small Resource utilization data of \sys{} implemented on GAP8 Processor}
\label{resource}
\centering
\scalebox{0.85}{
\resizebox{0.48\textwidth}{!}{%
\begin{tabular}{|c|c|c|c|}
\hline
\textbf{Resources}                                                          & \textbf{L1 Memory} & \textbf{L2 Memory} & \textbf{DRAM} \\ \hline
\textbf{\begin{tabular}[c]{@{}c@{}}Available for Use\\ (KB)\end{tabular}}   & 52.7               & 400                & 8000          \\ \hline
\textbf{\begin{tabular}[c]{@{}c@{}}TinyVQA\\ Utilization (KB)\end{tabular}} & 49 (93\%)            & 290 (73\%)           & 0             \\ \hline
\end{tabular}%
}}
\end{table}
\vspace{-2ex}

\begin{table}[]
\caption{\small Implementation Results of the proposed \sys{} and Comparison with Previous Work.}
\label{comparison}
\resizebox{0.48\textwidth}{!}{%
\begin{tabular}{|c|c|c|}
\hline
\textbf{Architecture}       & \textbf{TinyVQA (this work)} & \textbf{MobiVQA \cite{mobivqa}} \\ \hline
\textbf{Dataset}            & FloodNet-VQA \cite{floodnet}         & VQAv2 \cite{goyal2017making}            \\ \hline
\textbf{Modality Used}      & Image + Text     & Image + Text               \\ \hline
\textbf{Deployment Devices} & Gap8 Processor   & Nvidia TX2 Board           \\ \hline
\textbf{Frequency (MHz)}    & 175              & --                        \\ \hline
\textbf{Latency (ms)}       & 56               & 213                        \\ \hline
\textbf{Power (W)}         & 0.7              & --                        \\ \hline
\textbf{Energy (J)}        & 0.2              & 5.6                        \\ \hline
\end{tabular}%
}

\end{table}

\section{Conclusion}
The proposed TinyVQA system represents a significant advancement in the realm of tinyML, successfully addressing the challenges of integrating multiple data modalities into compact, low-power devices. Through meticulous evaluation on the FloodNet dataset, TinyVQA has proven its effectiveness by achieving 79.5\% accuracy in post-disaster scenarios, highlighting its real-world applicability. Furthermore, the deployment of TinyVQA on a power-efficient Crazyflie 2.0 drone equipped with an AI deck and GAP8 microprocessor exemplifies the system's operational proficiency. The tiny drone-based TinyVQA model, with its low latency of 56 ms and minimal power consumption of 0.7 W, underscores the potential of tinyML to revolutionize disaster assessment and response, opening new avenues for autonomous, intelligent systems in critical, resource-limited environments.

\section{ACKNOWLEDGMENT}

We acknowledge the support of the U.S. Army Grant No. W911NF21-20076.

\bibliographystyle{plain}
\bibliography{eehpc, tinyml}

\end{document}